\documentclass[letterpaper]{article} 
\usepackage{aaai25}  
\usepackage{times}  
\usepackage{helvet}  
\usepackage{courier}  
\usepackage[hyphens]{url}  
\usepackage{graphicx} 
\usepackage{natbib}  
\usepackage{caption} 
\usepackage{algorithm}
\usepackage{algorithmic}

\usepackage{newfloat}
\usepackage{listings}
\DeclareCaptionStyle{ruled}{labelfont=normalfont,labelsep=colon,strut=off} 
\floatstyle{ruled}
\newfloat{listing}{tb}{lst}{}
\floatname{listing}{Listing}

\usepackage{amsmath}
\usepackage{amsfonts}
\RequirePackage{xspace}

\makeatletter
\DeclareRobustCommand\onedot{\futurelet\@let@token\@onedot}
\def\@onedot{\ifx\@let@token.\else.\null\fi\xspace}

\makeatother
\title{RHanDS: Refining Malformed Hands for Generated Images with Decoupled Structure and Style Guidance}
\author{
    Chengrui Wang\equalcontrib\textsuperscript{\rm 1},
    Pengfei Liu\equalcontrib\textsuperscript{\rm 1, \rm 2}\thanks{Work done during the internship at Alibaba Group.},
    Min Zhou\textsuperscript{\rm 1}, 
    Ming Zeng\textsuperscript{\rm 2}\thanks{Corresponding author.},
    Xubin Li\textsuperscript{\rm 1},
    Tiezheng Ge\textsuperscript{\rm 1},
    Bo Zheng\textsuperscript{\rm 1}
}
\affiliations{
    \textsuperscript{\rm 1}Taobao \& Tmall Group of Alibaba\\
    \textsuperscript{\rm 2}Xiamen University\\
    \{wangchengrui.wcr, zhukong.lpf, yunqi.zm\}@taobao.com,\\
    zengming@xmu.edu.cn, \{lxb204722, tiezheng.gtz, bozheng\}@taobao.com
}

\usepackage{bibentry}

\begin{document}

\maketitle

\begin{abstract}
Although diffusion models can generate high-quality human images, their applications are limited by the instability in generating hands with correct structures. 
In this paper, we introduce RHanDS, a conditional diffusion-based framework designed to refine malformed hands by utilizing decoupled structure and style guidance.
The hand mesh reconstructed from the malformed hand offers structure guidance for correcting the structure of the hand, while the malformed hand itself provides style guidance for preserving the style of the hand.
To alleviate the mutual interference between style and structure guidance, we introduce a two-stage training strategy and build a series of multi-style hand datasets. 
In the first stage, we use paired hand images for training to ensure stylistic consistency in hand refining. 
In the second stage, various hand images generated based on human meshes are used for training, enabling the model to gain control over the hand structure.
Experimental results demonstrate that RHanDS can effectively refine hand structure while preserving consistency in hand style. 
\end{abstract}

%
\begin{links}
    \link{Code}{https://github.com/alimama-creative/RHanDS}
\end{links}

\begin{figure}[tb]
  \centering
  \includegraphics[width=\linewidth]{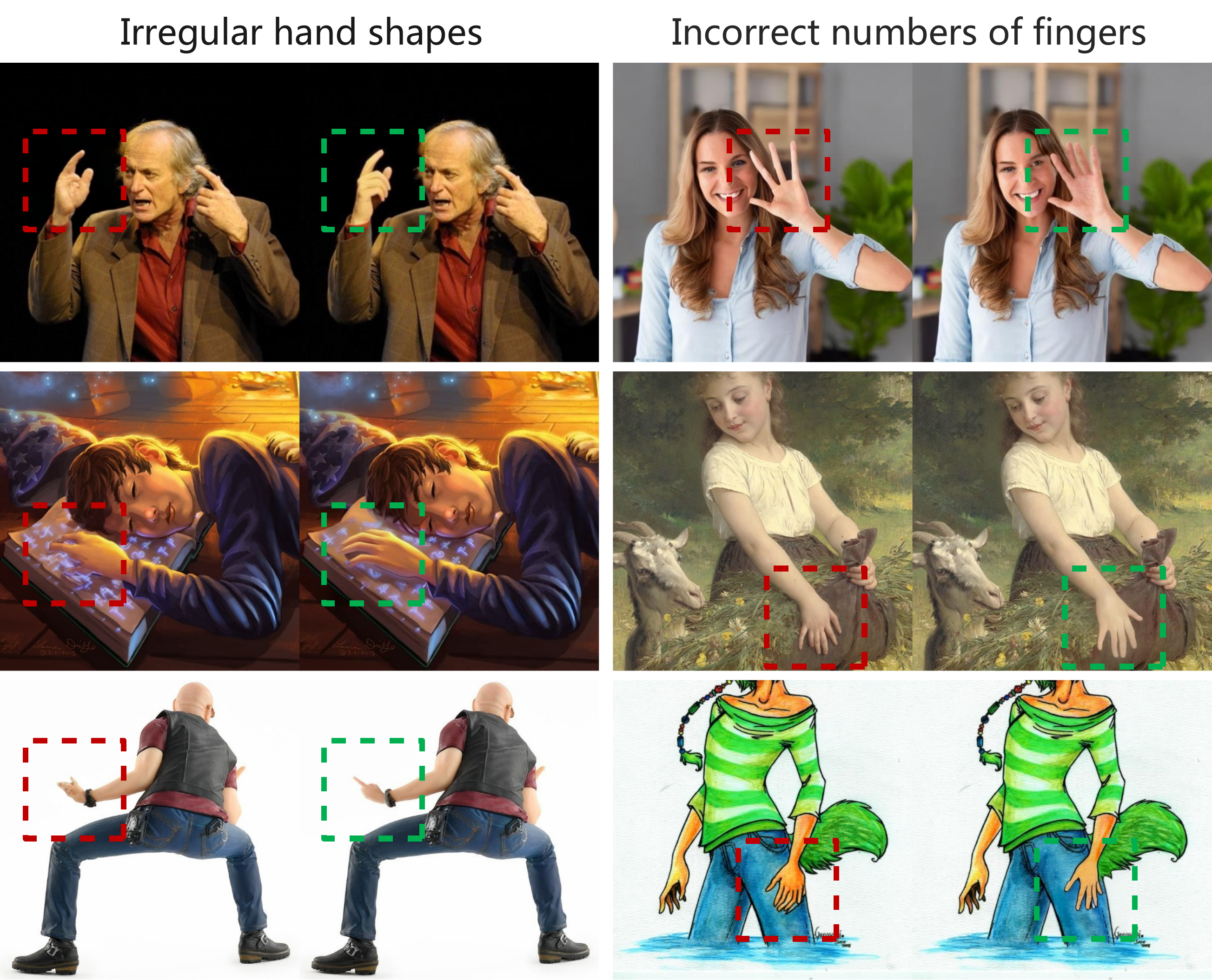}
  \caption{Examples of hands refined by our RHanDS (right in each pair) from the malformed hands (left in each pair).}
  \label{fig:overall}
\end{figure}

\begin{figure*}[tb]
\centering
    \includegraphics[width=1\textwidth]{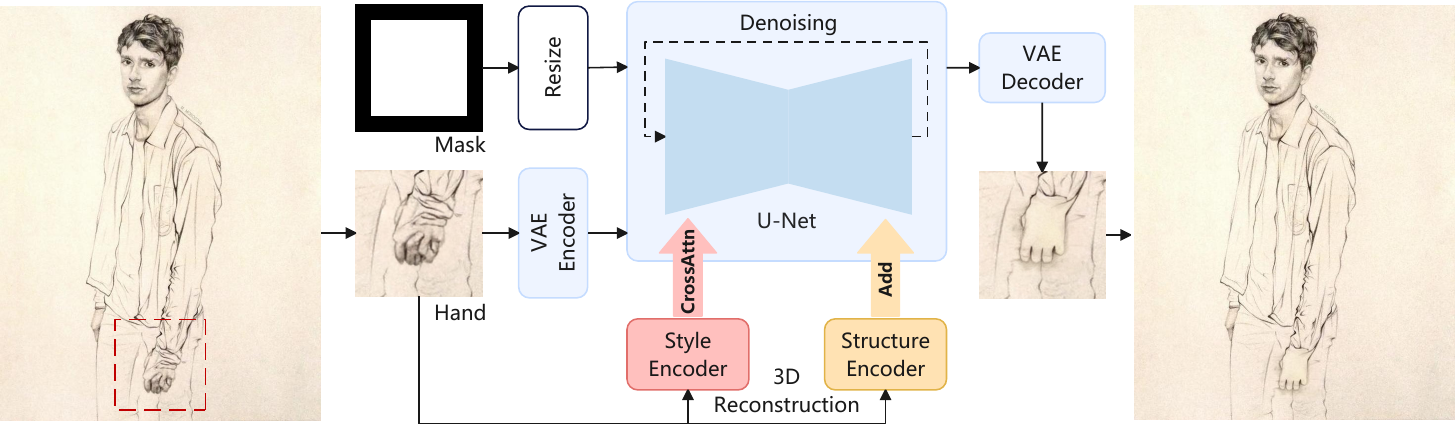}
\caption{
    The RHanDS framework we propose contains four modules: a VAE for projecting images into a latent space and reconstructing images from the latent, a conditional U-net for predicting the denoised variant during the denoising process, a style encoder to extract hand style from the malformed hand and map it into the U-net via cross-attention, and a structure encoder to utilize the hand mesh reconstructed from the malformed hand to guide the hand structure.
    In addition, to achieve a fully automatic process, a hand detection model and a 3D hand reconstruction model are required.
}
\label{fig:pipeline_main}
\end{figure*}

\section{Introduction}
Text-to-image diffusion models~\cite{nichol2022glide, rombach2022high, ramesh2022hierarchical, saharia2022photorealistic, podell2023sdxl} have exhibited the remarkable ability to synthesize visually stunning images based on textual prompts, representing a significant advancement in image generation.
Despite their impressive capabilities, current models still face challenges in effectively handling intricate structures like hands~\cite{podell2023sdxl,rombach2022high}. As illustrated in Figure~\ref{fig:overall}, models often produce malformed hands with irregular shapes or incorrect numbers of fingers, deviating from the realistic 3D shape and physical limitations of human hands.

To address the problem, \citeauthor{lu2024handrefiner} introduces a post-processing framework that employs an inpainting diffusion model to refine the malformed hands.
During the refining process, a 3D hand mesh is reconstructed based on the malformed hand image and rendered into a depth image to provide structure guidance for correcting the malformed hand through ControlNet~\cite{zhang2023adding}.
However, it is difficult to maintain the hand style as the refining process requires adding noise to the hand image.
This issue becomes particularly serious when addressing specific styles like sculptures or cartoons.
To enhance style consistency, it seems intuitive to utilize the image region of the original malformed hand.
However, extracting style information from the malformed hand inevitably introduces incorrect structural information~\cite{ye2023ip}.

In this paper, we propose a novel framework named RHanDS to refine malformed Hands with Decoupled Structure and Style guidance, which is adaptable to malformed hands of arbitrary styles.
As shown in Figure~\ref{fig:pipeline_main}, RHandDS utilizes diffusion models to repaint the hand region, effectively refining malformed hand structures while preserving the original hand style.
In detail, the 3D mesh reconstructed from the deformed hand is rendered into a depth image, which is then input into the structure encoder to provide structure guidance.
Meanwhile, the style encoder extracts the style embedding from the malformed hand  to offer style guidance. 
To alleviate the mutual interference between style and structure guidance, we introduce a two-stage training strategy and propose the corresponding multi-style hand datasets.

The first stage focuses on enabling the model to repaint hands in a specified style. To isolate hand styles from structural information, we collect a large number of realistic image pairs depicting the left and right hands of the same individual (i.e., images that share the same style but exhibit different structures). 
We then use one hand as a style reference to guide the generation of the other hand. In this stage, we only train the style encoder and U-net.

In the second stage, our objective shifts towards making the model generate hand images with additional structural control. 
Considering the limited style diversity in existing hand image datasets (with 3D meshes, data like these are called hand-mesh pairs for convenience in this paper), we propose a novel approach using the SMPL model and ControlNet to create a more diverse dataset. We initiate this process by generating hand meshes using the SMPL-H model and rendering the corresponding depth images. ControlNet is then utilized to generate hand images, controlled with depth images and prompts indicating various style categories.
In this stage, we train the structure encoder while keeping the other parameters frozen.

To verify the ability of hand refining, we use a standard diffusion model to regenerate the hand regions in the HumanArt dataset~\cite{ju2023human}, and propose a multi-style malformed hand dataset with mesh-image pairs for evaluation.
Both qualitative and quantitative experiments demonstrate the effectiveness and superiority of RHanDS.

In summary, the contributions of this work are as follows:
\begin{itemize}
    \item To refine malformed hand images, we propose a novel framework named RHandDS, which adopts decoupled structure and style guidance. By leveraging the direct and decoupled style guidance, our model can consistently maintains hand style while refining hand images across various styles.
    \item We propose a two-stage training strategy and create corresponding datasets, enabling the model to learn control over hand style and structure separately.
    \item Experiments demonstrate that RHanDS can effectively handle malformed hands across various styles and restore them to better structures.
\end{itemize}

\section{Related Work}
\subsection{Diffusion Models for Image Synthesis}
Recent advancements have been greatly promoted by text-to-image diffusion models.
Specifically, the technique of Latent Diffusion Models (LDM)~\cite{rombach2022high} conducts diffusion in a latent image space~\cite{van2017neural}, which largely reduces computational demands.
Stable Diffusion~\cite{sd15model} is one of the highlighted works of LDM, which utilizes a pre-trained language encoder like CLIP~\cite{radford2021learning} to encode the text prompt into latent space to guide the diffusion process.

Diverging from the text-to-image paradigm, image inpainting is tasked with synthesizing visually plausible content within the masked regions of the existing image, ensuring that the synthesized content maintains semantic coherence with the context of the unmasked regions.
By fine-tuning the pre-trained text-to-image Stable Diffusion model, the specialized variant known as Stable Diffusion Inpainting model~\cite{sd15inpaint} can synthesize content within the masked region that follows textual prompts.
However, fine-grained control over the structure and style of the inpainting contents cannot be achieved solely through text guidance. 

In order to provide fine-grained conditions during the generation process, different control methods have been explored in previous studies, and it has been demonstrated that an additional module can be effectively plugged into the existing diffusion models to guide the image generation.
ControlNet~\cite{zhang2023adding} and T2I-Adapter~\cite{mou2023t2i} use spatial conditions such as canny, depth images, and body poses to control the structure of the generated image.
To reduce the fine-tuning cost, Uni-ControlNet~\cite{zhao2024uni} allows for the simultaneous utilization of different conditions within one single model. 
Previous subject-driven approaches such as DreamBooth~\cite{ruiz2023dreambooth}, Textual Inversion~\cite{gal2022image}, LoRA~\cite{hu2021lora}, and Concept Sliders~\cite{gandikota2023concept} achieved stylization and customization by fine-tuning parameters.
In contrast, IP-Adapter~\cite{ye2023ip} leverages pre-trained image encoder~\cite{radford2021learning} to achieve the image prompt capability for diffusion models, generating images that resemble the reference images in content and style.

\subsection{Plausible Hand Generation}
Concept Sliders~\cite{gandikota2023concept} use parameter-efficient training method~\cite{hu2021lora} to learn better physical structure of hand by fine-tuning diffusion models with carefully selected data.
\citeauthor{ye2023affordance} synthesize hand-object interaction on existing images using the rough region of the palm and forearm as guidance.
However, the inherent complex structure of 3D human hands makes it difficult for these methods to generate correct hands stably.
Different from the methods that directly generate hands, some works~\cite{ye2023affordance, lu2024handrefiner, weng2023diffusion} introduce the structure information of the target hand into a model to prevent the model from generating malformed hands. 
HanDiffuser~\cite{ye2023affordance} encodes the 3D hand parameters into text embedding for achieving text-to-image generation with realistic hand appearances. 
Diffusion-HPC~\cite{weng2023diffusion} proposes a post-processing method that refines the generated malformed person through a conditional diffusion model with the help of a human depth image rendered based on the reconstructed human mesh.
HandRefiner~\cite{lu2024handrefiner} proposes a similar post-processing method more suitable for dealing with malformed hands.

Our RHanDS is a post-processing method that leverages 3D hand mesh as the pixel-level condition to control hand structure.
Compared to existing methods, RHanDS performs specifically on the hand region rather than the entire image, facilitating more precise refinement. 
Moreover, RHanDS enhances the perception of hand style, ensuring that the style of the refined hand seamlessly aligns with that of the original image for a coherent and authentic representation.

\section{Methodology}
In this section, we introduce the detail of our proposed framework RHanDS.
Given a generated image containing malformed human hands, we aim to correct the structure of the malformed hand while preserving the style of the generated image. 
As shown in Figure~\ref{fig:pipeline_main}, RHanDS mainly contains four modules: style encoder $\mathcal{E}_{style}$, structure encoder $\mathcal{E}_{struc}$, U-net $\epsilon_{\theta}$, and VAE encoder $\mathcal{E}$ and decoder $\mathcal{D}$.
The inputs of the framework include a cropped malformed hand image $x$, a structure reference $r_{struc}$, a style reference $r_{style}$, and a mask $mask$.
The malformed hand image $x$ is cropped out from the entire image for effective refining.
The structure reference $r_{struc}$ is a depth image rendered from a 3D hand model. The 3D hand model can be manually created using 3D tools~\cite{pose-editor} or automatically reconstructed from malformed hand, and the structure reference is encoded by the structure encoder $\mathcal{E}_{struc}$ to guide the hand structure.
The style reference $r_{style}$ is the malformed hand itself and is encoded by the style encoder $\mathcal{E}_{style}$ to guide the hand style.
The binary mask $mask$ indicates the malformed hand region that needs to be repainted.
The training process is decomposed into two stages to decouple the structure and style guidance. 
In the first stage, the U-net and style encoder are trained using Multi-Style Paired Hand Dataset.
In the second stage, the structure encoder is trained using the Multi-Style Hand-Mesh Dataset.

\subsection{Preliminaries}\label{prelim}
RHanDS is built on stable diffusion, which efficiently performs the diffusion process in latent space rather than pixel space.
Specifically, given an image $x_0$ in RGB space, the encoder $\mathcal{E}$ first encodes the image into a latent representation $z_0=\mathcal{E}(x_0)$. 
During the forward process, normally-distributed noise $\epsilon_t$ is added into the latent $z_0$ to obtain the noisy latent $z_t$.
During the reverse process, the stable diffusion model implements denoising U-net $\epsilon_{\theta}$ as the backbone to predict the noise $\epsilon_t$ with noisy latent $z_{t}$ and current timestep $t$.
The simple loss function can be written as 
\begin{align}
    \mathcal{L} = \mathbb{E}_{z_0, \epsilon \sim \mathcal{N}(0,1), t} [|| \epsilon - \epsilon_{\theta}(z_t, t)||^{2}_{2}].
\end{align}

In this work, RHanDS employs a U-net with five additional input channels.
Specifically, four channels are allocated for the encoded masked-image $\hat{z}_0$, and the remaining single channel is allocated for the resized mask $mask$. 
During reverse process, the U-net predicts the noise through $\epsilon_t = \epsilon_{\theta}(z_t, \hat{z}_0, mask, t)$, where $\hat{z}_0$ and $mask$ keep unchanged.
After the reverse process, the decoder $\mathcal{D}$ reconstructs the hand image from the latent. 

\begin{figure}[t]
    \centering
    \includegraphics[width=\linewidth]{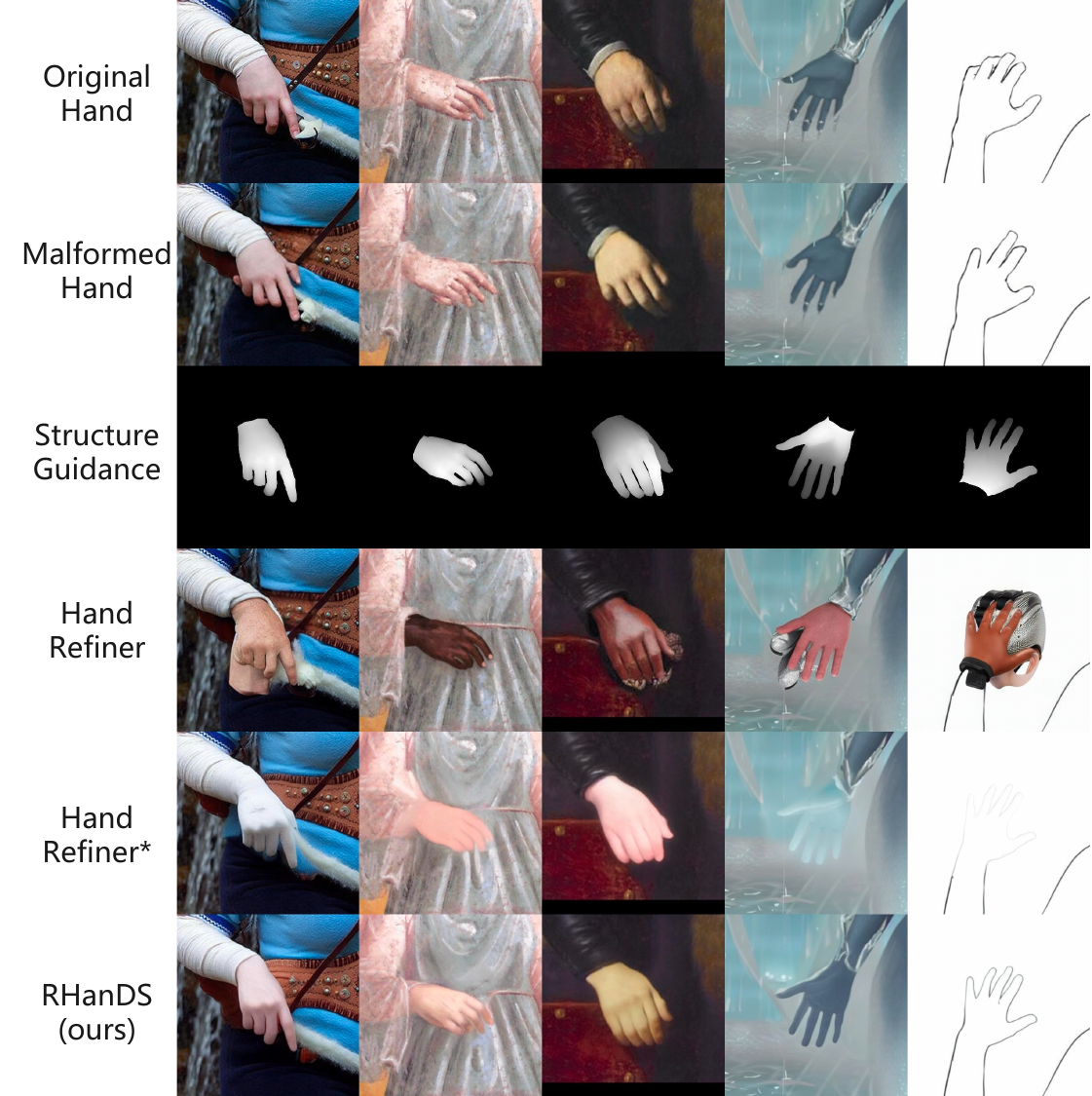}
    \caption{
        The visual comparison of RHanDS with other methods on different styles of malformed hands. 
        The malformed hands are generated based on the original hands, and the structure guidances are reconstructed from the original hands.
    }
    \label{fig:method_comparison}
\end{figure}

\subsection{The First Stage: Style Guidance}

\begin{figure}[tb]
  \centering
  \includegraphics[width=\linewidth]{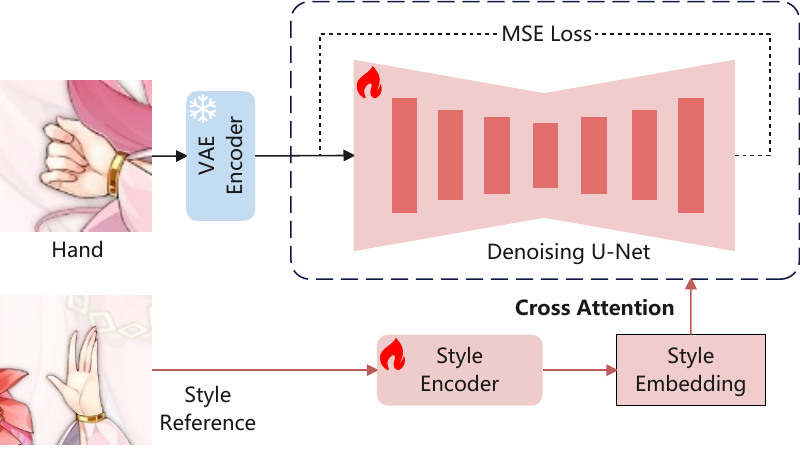}
  \caption{The first stage. In this stage, U-net and style encoder are trained using Multi-Style Paired Hand Dataset for style guidance.}
  \label{fig:stage1}
\end{figure}

As shown in Figure~\ref{fig:method_comparison}, the existing solutions are limited in perceiving hand style and exhibit poor generalization ability in images with anime, oil painting, and other styles.
To solve the problem, we propose to encode the hand style into U-net, enabling the model to refine hands $x$ based on style reference $r_{style}$.
Similar to the approach used in IP-Adapter~\cite{ye2023ip}, we use CLIP~\cite{radford2021learning} image encoder to extract the global image embedding from $r_{style}$, and then use a linear projection network to transform the image embedding into a sequence of features to obtain the style embedding $c_{style}$.
Overall, the style encoder $\mathcal{E}_{style}$ consists of a CLIP image encoder and a linear projection network.
The style embedding $c_{style}$ that can be formulated as follows:

\begin{align}
c_{style} = \mathcal{E}_{style}(r_{style})
\end{align}
Since the style embedding shares the same dimensions as the original text embedding in U-net, we directly substitute the text embedding with the style embedding.
As shown in Figure~\ref{fig:stage1}, in this stage, the U-net and the style encoder are trained on multi-style paired hands.
We randomly choose a hand from the hand pairs and add noise $\epsilon$ to obtain the noisy latent $z_t$, and use the other hand to supply style embedding $c_{style}$ to prevent the model from learning hand structure from style guidance.
The loss function is as follows:
\begin{align}
    \mathcal{L} = \mathbb{E}_{z_t, \epsilon, t} [|| \epsilon - \epsilon_{\theta}(z_t, c_{style}, mask, t)||^{2}_{2}],
\end{align}

To build the multi-style paired hand dataset that meets the specified requirements, we collect character images of varying styles, and crop images of two hands from the same character, ensuring that each pair exhibits identical styles but different gestures.
The details of the dataset can be found in Section~\ref{sec:d1}.

\begin{figure}[tb]
  \centering
  \includegraphics[width=\linewidth]{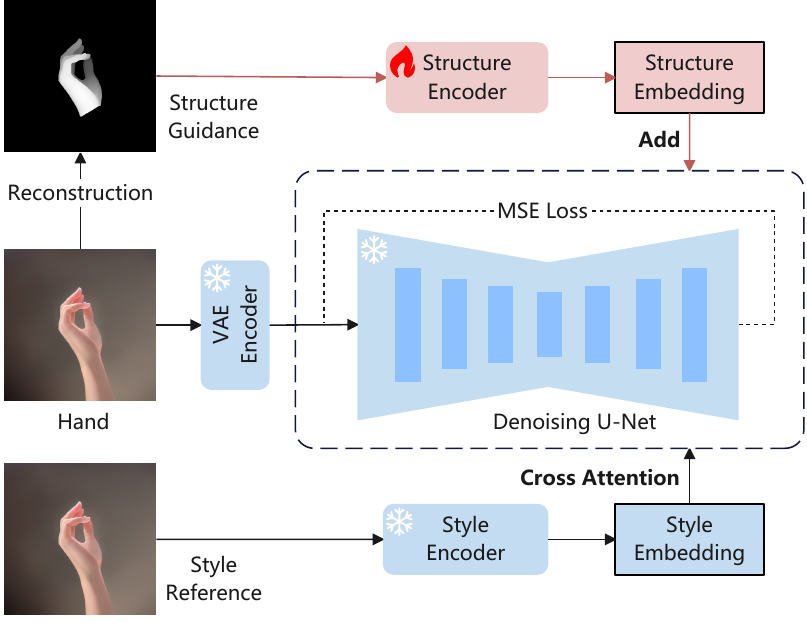}
  \caption{The second stage. In this stage, the structure encoder is trained using Multi-Style Hand-Mesh Dataset for structure guidance.}
  \label{fig:stage2}
\end{figure}

\subsection{The Second Stage: Structure Guidance}
Under style guidance, our framework can repaints hands while preserving their original style.
However, as shown in Figure~\ref{fig:modules}, it is hard to correct the hand structure stably without explicit structure guidance due to the inherent complexity of hand structure.
To address the problem, we introduce the representation of the correct hand structure by estimating the MANO-based~\cite{romero2022embodied} hand mesh from the malformed hand.
For malformed hands that are difficult to reconstruct automatically, manual recontruction can be performed using existing tools~\cite{pose-editor}. 
Following~\citeauthor{lu2024handrefiner}, we render the hand mesh into a depth image, serving as the structure reference $r_{struc}$, to guide the correction of hand structure.

We use the structure encoder $\mathcal{E}_{struc}$ to encode the structure reference $r_{struc}$ into U-net for structure guidance, where $\mathcal{E}_{struc}$ has the same structure as ControlNet~\cite{zhang2023adding}.
Moreover, we feed a learnable embedding $c_{l}$ instead of text embedding into the cross-attention layers of $\mathcal{E}_{struc}$.
The learnable embedding shares the same dimension as the text embedding and is integrated into the diffusion model similarly to how the text embedding is.
Specially, the input of $\mathcal{E}_{struc}$ is consist of structure reference $r_{struc}$, noisy latent $z_{t}$ and learnable embedding $c_{l}$. The output of $\mathcal{E}_{struc}$ is the structure embedding $c_{struc}$ that can be formulated as follows:
\begin{align}
c_{struc} = \mathcal{E}_{struc}(r_{struc}, z_{t}, c_{l})
\end{align}
We feed the structure embedding $c_{struc}^{i}$ into all the outputs $out^{i}$ of the middle block and decoder blocks of U-net through a $1 \times 1$ convolution layer $\mathbb{Z}^{i}$ with structure weight $w$:
\begin{align}
out^{i} = out^{i} + w \cdot \mathbb{Z}^{i}(c_{struc}^{i})
\end{align}

As shown in Figure~\ref{fig:stage2}, in this stage, we freeze the U-net and the style encoder to maintain the capability of style guidance and train the structure encoder on the multi-style hand-mesh dataset.
We add noise $\epsilon$ into the hand image to obtain the noisy latent $z_t$ and use the structure encoder to obtain the $c_{struc}$ from the corresponding hand mesh.
\begin{align}
    \mathcal{L} = \mathbb{E}_{z_t, \epsilon, t} [|| \epsilon - \epsilon_{\theta}(z_t, c_{style}, c_{struc}, mask, t)||^{2}_{2}].
\end{align}

However, even if we only fine-tune the structure encoder, the performance of style guidance degrades when training with style-limited datasets datasets~\cite{zb2017hand,chen2022mobrecon}.
Therefore, based on the depth-controlled text-to-image generation pipeline, we use stable diffusion and ControlNet to create a multi-style hand-mesh dataset by using the depth image rendered from random hand meshes.
The details of this dataset can be found in Section~\ref{sec:d2}.

\section{Multi-Style Dataset}
We have introduced a hand-refining model that decouples style and structure guidance, along with a two-stage training strategy. In this section, we will introduce a series of datasets purposefully constructed for this method. 
They encompass various data types and styles of hands. Detailed information about these datasets is provided below.

\subsection{Multi-Style Paired Hand Dataset}\label{sec:d1}
During the first-stage training, to enable the model to refine hands under style guidance while preventing the leakage of hand structure, we propose a dataset consisting of hand pairs in various styles, each with the same style but different gestures.
We find these pairs by detecting the two hands of the same person. Specifically, we employ the YOLOv8~\cite{yolov8} to detect human hands and use mmpose~\cite{mmpose2020} to detect human pose, which allows us to filter out hands belonging to the same person. 
This dataset includes 517,096 hand pairs from natural human images with various textures, and 2,423 hand pairs from anime images.

\subsection{Multi-Style Hand-Mesh Dataset}\label{sec:d2}
During the second-stage training, to prevent the model from overfitting to the generation of hands in a single style, we first utilize the SMPL model with MANO (SMPL-H)~\cite{loper2023smpl, MANO:SIGGRAPHASIA:2017} to create hand meshes and get corresponding depth images. Specifically, we randomly generated the body pose parameters of the SMPL model and hand pose parameters of the MANO model, then combined them to form a complete human model. We cropped out the arm and hand submesh and rendered them into depth images. Then, we employ the depth-conditioned ControlNet to generate multi-style hand images with text prompts indicating different style categories.
We reconstruct the hand mesh from the generated hand images and calculate the metric $\textrm{MPJPE}(K, K') = \mathbb{E}||K-K'||$ to filter out poorly structured hands, where $K$, $K'$ are the 2D projection coordinates of the hand joints, corresponding to the MANO hand mesh and the reconstructed hand mesh.
Ultimately, this dataset includes 7 different style categories, with 8,000 hand-mesh pairs for each style.

\subsection{Multi-Style Malformed Hand-Mesh Dataset}\label{sec:d3}
To evaluate the model's ability to handle multi-style hands, we extract hand images from the HumanArt~\cite{ju2023human} dataset and use SDEdit~\cite{DBLP:conf/iclr/MengHSSWZE22} to regenerate the hand regions as corresponding malformed hand images.
For evaluation, the style reference is the malformed hand itself, while the structure reference is reconstructed from the original hands using~\cite{chen2022mobrecon}.
After manually filtering out the data with reconstruction failures, we collect 1440 image pairs in total, consisting of 1 natural style and 13 artificial styles.

\section{Experiments}
\subsection{Implementation Details}
In order to utilize the generation capability of existing model, the VAE and U-net are initialized from Stable Diffusion Inpainting v1.5, and the structure encoder is initialized from pretrained depth ControlNet.
CLIP ViT-H/14~\cite{ilharco_gabriel} is adopted as the style encoder, and the linear projection network is randomly initialized. 

For the first stage, we use our Multi-Style Paired Hand Dataset to train U-net and style encoder for 15k iterations with a learning rate of 1e-5 and with a total batch size of 256 on 8 NVIDIA-A100 GPUs.
The default token length for the style encoder is set to 8.
The hand images $x$ are resized to $512 \times 512$, and the style references $r_{style}$ are resized to $224 \times 224$ with random horizontal and vertical flipping.

For the second stage, we use the combination of Static Gesture Dataset (SGD)~\cite{static-gesture-dataset} and our Multi-Style Hand-Mesh Dataset to train structure encoder for 15k iterations with a learning rate of 2e-5 and with a total batch size of 256 on 8 NVIDIA-A100 GPUs.
The weight $w$ for the structure encoder is set to 1 during training.
The structure references $r_{struc}$ are resized to $512 \times 512$ and the depth values of the hand are normalized into [0.2, 1.0].

Additionally, the mask $mask$ we use to indicate the hand region covers the middle 9/16 of the entire image.
In order to learn style guidance from the style encoder rather than the remaining skins in the image, we apply a strategy to randomly expand the region indicated by the mask to force the model to repaint all the skin in the image.
Meanwhile, we randomly replace the style embedding with a learnable style embedding with a probability of 0.1 and set the mask to the entire image with a probability of 0.5.
We adopt noise offset~\cite{noiseoffset} and min-SNR strategy~\cite{DBLP:conf/iccv/HangGLB00GG23} for training. 

During inference, we add the maximum noise (strength = 1.0) to image $x$ and adopt DDPM~\cite{DBLP:conf/nips/HoJA20} sampler for denoising.
We use structure weight $w$ = 0.6, denoising step $steps$ = 25 as default.
The refining process takes approximately 3 seconds and requires 4GB GPU memory.

\begin{table}[t]
\centering
\begin{tabular}{lccc}
\hline
Method  & FID$~\downarrow$ & MPJPE $~\downarrow$ & Conf.$~\uparrow$ \\ \hline
\multicolumn{4}{c}{Text2Image Dataset} \\ 
Stable Diffusion\dag & 77.60& -     & 0.93     \\
HandRefiner\dag      & 74.12& -     & 0.94     \\
RHanDS\dag & \textbf{73.63}& -     & 0.94     \\ \hline
\multicolumn{4}{c}{Image2Image Dataset}\\
HandRefiner\dag      & 13.97  & 7.87   & 0.97     \\
RHanDS\dag  & \textbf{13.54}  & \textbf{6.89}     & 0.97    \\
HandRefiner     &  33.84    & 12.02        & 0.94 \\
RHanDS & \textbf{22.18}& \textbf{9.86}    & \textbf{0.96}     \\ \hline
\end{tabular}
\caption{
    Quantitative comparison of the proposed RHanDS with other methods on datasets~\cite{lu2024handrefiner}. 
    The method with \dag\ represents calculating metrics on the entire image, while the method without \dag\ represents calculating metrics only on the hand region.
}
\label{tab:compare_hr}
\end{table}

\subsection{Evaluation Metrics}
We use several metrics to evaluate the performances: 
(1) Fr$\acute{e}$chet Inception Distance~\cite{heusel2017gans} (FID) focuses on the overall distribution statistics of the refined images and the Ground Truth, 
(2) Style loss\cite{gatys2016image} evaluates the style consistency between the refined images and the Ground Truth (the loss we reported is multiplied by 100), 
(3) MPJPE~\cite{ionescu2013human3} between reconstructed 2D hand poses measures the structure consistency, 
(4) Keypoint detection confidence scores (Conf.) of a hand detector~\cite{lugaresi2019mediapipe, zhang2020mediapipe} is used to indicate the plausibility of generated hands.
The FID and style loss are calculated between the refined hands and the original hands.
The MPJPE is calculated between the reconstructed hand mesh and the structure reference, where the hand mesh is reconstructed from the refined hand using MeshGraphormer~\cite{lin2021mesh} for test datasets proposed by HandRefiner and using MobRecon~\cite{chen2022mobrecon} for our multi-style multi-style malformed hand-mesh test dataset. 
By default, MPJPE is calculated based on the resolution of $512 \times 512$.

\begin{table*}[th]
    \centering
    \begin{tabular}{l|cc|cccc}
    \hline
    Method & Style Encoder  & Structure Encoder & MPJPE\ $\downarrow$ & Conf.\ $\uparrow$ & FID\ $\downarrow$  & Style Loss\ $\downarrow$ \\ 
    \hline
    Malformed Dataset   & -        & - & 40.29        & 0.86 & 23.84       & 2.77    \\
    HandRefiner& -  & - & 33.27    & 0.88 & 38.52       & 10.13   \\ 
    HandRefiner* & -  & - & 29.60    & 0.89 & 34.12       & 5.61   \\ 
    \hline
     
    RHanDS      & stage-2 & stage-2  & 26.29        & 0.89  & 47.76       & 7.11    \\
    RHanDS      & stage-1*& stage-2  & 29.48        & 0.88& 33.68       & 5.35    \\
    RHanDS      & stage-1 & stage-2* & \textbf{26.17} & \textbf{0.91}     & 33.56       & 5.23    \\
    \textbf{RHanDS}      & stage-1 & stage-2  & 27.23        & 0.89 & \textbf{32.59}& \textbf{5.04}\\
    \hline
    \end{tabular}
    \caption{
    Quantitative comparison of the proposed RHanDS with other methods on the Multi-Style Malformed Hand-Mesh Dataset. Stage-1* denotes using the hand image itself as a style reference for the first stage training, and stage-2* denotes using only the SGD dataset for the second stage training.
    }
    \label{tab:comparison}
\end{table*}

\begin{figure}[tb]
    \centering
    \includegraphics[width=0.9\linewidth]{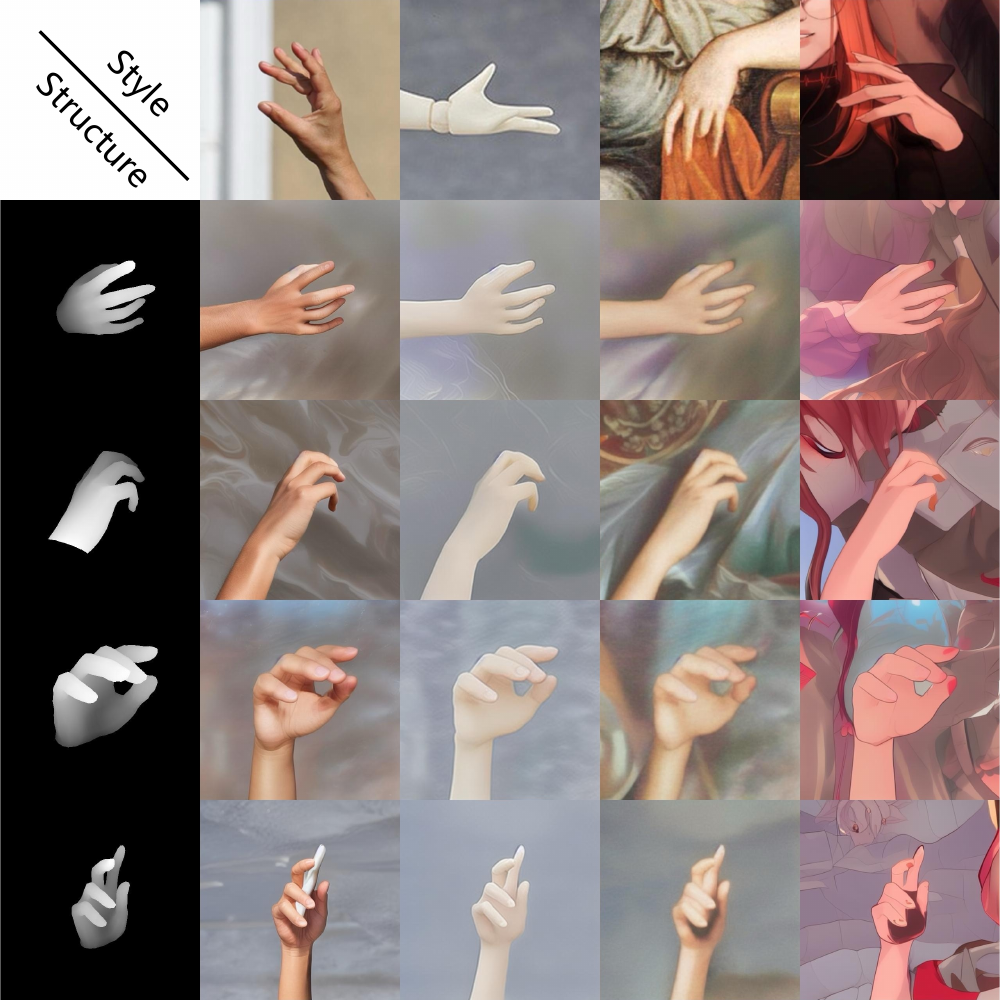}
    \caption{Visualization of hands generated under the guidance of different style and structure references.}
    \label{fig:style_structure}
\end{figure}

\subsection{Results and Comparison}

\textbf{Comparison between different methods on HandRefiner Dataset. }
We evaluate our method on two datasets proposed by HandRefiner.
The Text2Image Dataset includes 12K images generated with the text descriptions from HAGRID~\cite{kapitanov2024hagrid}, and the Image2Image Dataset includes 2K images sampled from HAGRID.
Following the setting in HandRefiner, for each image, we extract the hand region and refine the hand, then paste it back to the original image and calculate metrics on the entire image.
Meanwhile, we found that calculating metrics on the hand region extracted using a fixed strategy can eliminate the influence of hand proportion and image size on the evaluation metrics.
Therefore, we also report the metrics calculated on the hand region. 
As shown in Table~\ref{tab:compare_hr}, our RHanDS outperform HandRefiner on both datasets.

\textbf{Style-structure cross reference.}
To verify the robustness and generalization of RHanDS, we conducted experiments with various styles across different structure references. 
During inference, we mask the entire image to achieve the more intuitive result in Figure~\ref{fig:style_structure}.
Since the style and structure guidance are decoupled during training, we can generate hands with specified structures under any style reference without worrying about structure leakage.

\textbf{Comparison between different methods on Multi-Style Malformed Hand-Mesh Dataset. }
We compare with HandRefiner on our multi-style malformed hand-mesh dataset to evaluate the generalization ability to various styles. For fair comparison, we also reimplement and train a variant version of handrefiner on our multi-style dataset, which is denoted as HandRefiner*.
As shown in Figure~\ref{fig:method_comparison}, the hands generated by HandRefiner are monotonous in terms of style.
After training with our multi-style dataset, HandRefiner* can generate hands of various styles while still struggling to generate hands with the consistent color of the original malformed hands. 
Through decoupled style and structure guidance, our RHanDS refines hand structure while maintaining consistency with the original malformed hand style.
Furthermore, the quantitative experiments in Table~\ref{tab:comparison} show that our RHanDS outperforms HandRefiner and HandRefiner* in all the evaluation metrics in terms of style and structure.

\begin{figure}[t]
    \centering
    \includegraphics[width=0.95\linewidth]{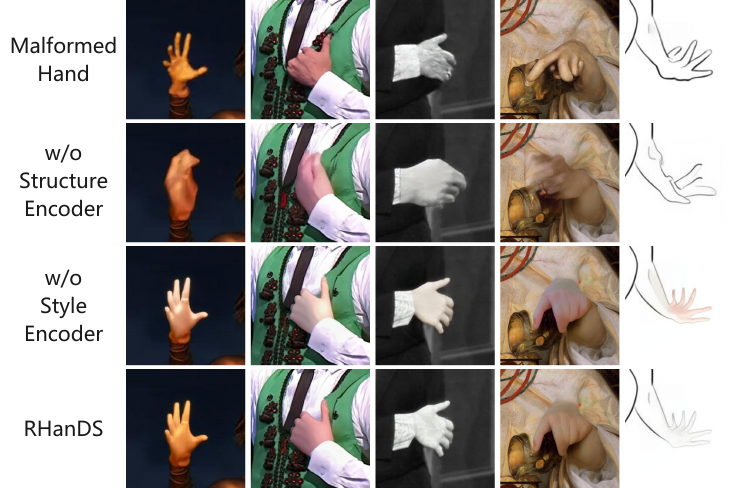}
    \caption{The visual comparison of different modules.}
    \label{fig:modules}
\end{figure}

\subsection{Ablation Study}

In this subsection, we will validate the effectiveness of the two-stage training strategy and dataset construction through quantitative experiments. Additionally, we will analyze the roles of the style and structure modules within the network.

\textbf{Effect of two-stage training.}
Without the first-stage training, we train the entire model on the multi-style hand-mesh dataset with a single hand as both hand image $x$ and style reference $r_{ref}$.
As shown in Table~\ref{tab:comparison}, this model can achieve performance comparable to the two-stage training model in terms of MPJPE and  Conf., but the FID has increased by 15.17, and the style loss has increased by 2.07.
Additional experiment shows that compared with the the model without structure encoder (FID 33.33, Style Loss 5.50), the FID is increased by 14.43 and the style loss is increased by 1.61.
These demonstrate that the first-stage training is beneficial for helping the model perceive hand style from style reference.

\textbf{Effect of Paired Hand Dataset.}
To clarify the impact of hand structure information leaked from the style encoder on hand refinement, we use the unpaired hands for the first stage training, where the style reference $r_{ref}$ uses the same image as the hand image $x$.
As shown in Table~\ref{tab:comparison}, the model trained with paired hands achieves better performance on all metrics, especially on the metrics related to hand structure. 

\textbf{Effect of Multi-Style Hand-Mesh Dataset.}
To clarify the role of multi-style hand-mesh data in two-stage training, we conduct an experiment using only the SGD dataset for the second-stage training.
As shown in Table~\ref{tab:comparison}, compared with using both SGD and the multi-style hand-mesh dataset, the model trained with the SGD dataset outperformed in terms of MPJPE and Conf. by 1.06 and 0.02, respectively. 
The reason for better MPJPE and  Conf. is that, unlike the hand images generated using stable diffusion in the multi-style hand-mesh dataset, the hand images rendered from mesh in SGD are more stable in hand structure.
However, due to the low style diversity of hand data in SGD, the FID increased by 0.97, and the style loss increased by 0.19.

\textbf{Effects of different modules. }
In RHanDS, we use style encoder and structure encoder to guide the style and structure of the hands. 
To illustrate the effects of these two modules, we separately shield the structure encoder and the style encoder during inference.
Specifically, we set the structure weight $w$ to 0 to shield the structure encoder and use the learnable style embedding rather than extracting embedding from the style reference to shield the style encoder.
Qualitative analysis in Figure~\ref{fig:modules} shows that the structure of the hand tends to be malformed without the structure encoder, and the style of the hand is easily changed without the style encoder.

\begin{table}[t]
\centering
\begin{tabular}{lcccc}
\hline
$w$ & MPJPE\ $\downarrow$ & Conf.\ $\uparrow$ & FID\ $\downarrow$  & Style Loss\  \ $\downarrow$  \\
\hline
0.0 & 67.19 & 0.833 &  33.33  & 5.50 \\
0.2 & 48.32 & 0.867 & 31.50 & 4.96 \\
0.4 & 31.50 & 0.895 & \textbf{31.20} & \textbf{4.71} \\
0.6 & \textbf{27.23}& \textbf{0.896} & 32.59 & 5.04 \\
0.8 & 27.29 & 0.891 & 34.46 & 5.62 \\
1.0 & 29.73 & 0.881 & 37.46 & 6.22\\
\hline
\end{tabular}
\caption{Ablation study on the impact of structure weight $w$.}
\label{tab:comparison2}
\end{table}

\textbf{Structure weights. }
The structure weight $w$ is adjustable during inference.
As shown in Table~\ref{tab:comparison2},  the optimal value of structure weight is almost in the range of 0.4 to 0.6.
Specifically, we choose the setting of w = 0.6 as optimal.
This setting achieves the best MPJPE and Conf., which is important in controlling hand structure.

\section{Conclusion}
In this paper, we introduce a novel hand refining method RHanDS.
It decouples the hand refining process into style and structure guidance and employs a two-stage training strategy to fully leverage existing data, improve the structure of repainted hands, and preserve the style of the original image. Furthermore, three datasets designed specifically for this task are built to facilitate research on multi-style hand refining.
Qualitative and quantitative experiments demonstrate our method's superiority in adapting to images across various styles compared with existing methods.

\section*{Acknowledgements}
This work is partially supported by National Natural Science Foundation (Grant No. 62072382), Yango Charitable Foundation, and Alibaba Group through Alibaba Research Intern Program.
\bibliography{aaai25}

\end{document}